%% file: main.tex
\documentclass[letterpaper, 10pt, conference]{ieeeconf/ieeeconf}    % Use this line for a4 paper

\makeatletter
\let\NAT@parse\undefined
\makeatother

\usepackage[numbers,sectionbib,sort&compress]{natbib}

\usepackage{gensymb}
\usepackage{graphicx}
\usepackage{amsmath}
\usepackage{algorithm}
\usepackage[noend]{algpseudocode}
\usepackage{subcaption}
\usepackage[font=small]{caption}
\usepackage{sidecap} \sidecaptionvpos{figure}{c}
\usepackage{amsfonts}
\usepackage{float}

\usepackage{algorithmicx}
\algdef{SE}[DOWHILE]{Do}{doWhile}{\algorithmicdo}[1]{\algorithmicwhile\ #1}%

\usepackage{hyperref}
\hypersetup{colorlinks,breaklinks,
linkcolor=[rgb]{0.5,0.,0.},
citecolor=[rgb]{0.000,0.427,0.173},
urlcolor=[rgb]{0.031,0.318,0.612}}

\usepackage[printonlyused,withpage,nolist,nohyperlinks]{acronym}
\input{common_acronyms.tex}

\usepackage[normalem]{ulem}

\IEEEoverridecommandlockouts                           % This command is only needed if
\title{\LARGE \bf
Free Energy Principle for State and Input Estimation of a Quadcopter Flying in Wind
}

\author{$\text{Fred Bos \textsuperscript{1}}$, $\text{Ajith Anil Meera \textsuperscript{2}} $, $\text{Dennis Benders \textsuperscript{3}}$ and $\text{Martijn Wisse \textsuperscript{4}}$ \thanks{All the authors are with
the Cognitive Robotics department at TU Delft, The Netherlands. Corresponding author: \texttt{ajitham1994@gmail.com}.}%
}

\begin{document}

\maketitle
\thispagestyle{empty}
\pagestyle{empty}

%%%%%%%%%%%%%%%%%%%%%%%%%%%%%%%%%%%%%%%%%%%%%%
\input{sections/Abstract}
\input{sections/Introduction}

\input{sections/Related_work}
\input{sections/Problem_statement}
\input{sections/Preliminaries}
\input{sections/Experimental_design}

\input{sections/Results}

\section{CONCLUSIONS AND FUTURE WORK} \label{S:conclusion}
The FEP based perception scheme called DEM, has recently been reformulated into a simulataneous state and input observer for LTI systems under colored noise. With this paper, we propose an experimental design to validate the DEM observer on real robots. Through a series of quadrotor experiments under wind conditions, we show that the DEM based observer outperforms other benchmarks like KF, SMIKF and SA for state estimation with minimum estimation errors. We show that DEM's input estimation shows similar performance compared to classical input observers like UIO. With this paper, we provide the first experimental validation for the use of generalized coordinates to deal with colored noise during state and input estimation on real robots. We further demonstrate the unique capability of DEM to balance between accuracy and complexity during state and input estimation.
% The paper provided the first experimental validation for the use of the dynamic expectation maximisation algorithm in robotics. DEM leverages the coloured nature in system disturbances by making use of generalized coordinates. DEM has shown to outperform conventional methods in state estimation, such as the Kalman filter, SMIKF and state augmentation, while at the same time estimating the systems inputs. 
The main challenge of the DEM observer is the need to know the noise precision and noise smoothness \textit{a priori}. We intend to extend this work for simultaneous noise precision and smoothness estimation in future.

\section{ACKNOWLEDGEMENT}
We would like to thank Prof. Tamás Keviczky for letting us use the lab facilities in the midst of corona crisis.

\bibliographystyle{IEEEtranN}
\footnotesize
\bibliography{references/IEEEabrv,references/2019-anil-meera}

\end{document}

%% file: common_acronyms.tex
\begin{acronym}

\acro{2D}{two-dimensional}

\acro{3D}{three-dimensional}

\acro{AHRS}{attitude and heading reference system}

\acro{AUV}{autonomous underwater vehicle}

\acro{CPP}{Chinese Postman Problem}

\acro{DoF}{degree of freedom}
\acrodefplural{DoF}[DoFs]{degrees of freedom}

\acro{DVL}{Doppler velocity log}

\acro{FSM}{finite state machine}

\acro{IMU}{inertial measurement unit}

\acro{LBL}{Long Baseline}

\acro{MCM}{mine countermeasures}

\acro{MDP}{Markov decision process}
\acrodefplural{MDP}[MDPs]{Markov decision processes}

\acro{POMDP}{partially observable Markov decision process}
\acrodefplural{POMDP}[POMDPs]{partially observable Markov decision processes}

\acro{PRM}{Probabilistic Roadmap}
\acrodefplural{PRM}[PRM]{Probabilistic Roadmaps}

\acro{ROI}{region of interest}
\acrodefplural{ROI}[ROIs]{regions of interest}

\acro{ROS}{Robot Operating System}

\acro{ROV}{remotely operated vehicle}

\acro{RRT}{Rapidly-exploring Random Tree}
\acrodefplural{RRT}[RRTs]{Rapidly-exploring Random Trees}

\acro{SLAM}{simultaneous localization and mapping}

\acro{SSE}{sum of squared errors}

\acro{STOMP}{Stochastic Trajectory Optimization for Motion Planning}

\acro{TRN}{Terrain-Relative Navigation}

\acro{UAV}{unmanned aerial vehicle}

\acro{USBL}{Ultra-Short Baseline}

\acro{IPP}{informative path planning}

\acro{FoV}{field of view}
\acrodefplural{FoV}[FoVs]{fields of view}

\acro{CDF}{cumulative distribution function}

\acro{ML}{maximum likelihood}

\end{acronym}

%% file: sections/Abstract.tex
\begin{abstract}
% The Free energy principle from neuroscience provides a brain inspired inference scheme for perception and action through a data-driven model inversion algorithm called Dynamic Expectation Maximization (DEM). The use of generalized coordinates under the free energy principle framework enables DEM to leverage the coloured nature of noises to provide accurate state and input estimation. This research tests the capabilities of DEM on a quadcopter influenced by coloured noise in the form of wind. The result is the experimental confirmation of the usefulness of DEM in robotics and an observer solution for state and input estimation for systems influenced by coloured noise. The superior performance of DEM is demonstrated in comparison to other benchmarks such as the Kalman filter, SMIKF and state augmentation, with minimal errors in state prediction. The paper concludes by highlighting DEM input estimation in comparison to an unknown input observer. 

The free energy principle from neuroscience provides a brain-inspired perception scheme through a data-driven model learning algorithm called Dynamic Expectation Maximization (DEM). This paper aims at introducing an experimental design to provide the first experimental confirmation of the usefulness of DEM as a state and input estimator for real robots. Through a series of quadcopter flight experiments under unmodelled wind dynamics, we prove that DEM can leverage the information from colored noise for accurate state and input estimation through the use of generalized coordinates. We demonstrate the superior performance of DEM for state estimation under colored noise with respect to other benchmarks like State Augmentation, SMIKF and Kalman Filtering through its minimal estimation error. We demonstrate the similarities in the performance of DEM and Unknown Input Observer (UIO) for input estimation. The paper concludes by showing the influence of prior beliefs in shaping the accuracy-complexity trade-off during DEM's estimation.

\end{abstract}

%% file: sections/Introduction.tex
\section{INTRODUCTION} \label{S:introduction}

%Intro about how impactful the state and input estimation for drones is. Why is modeling coloured noises important?

%Give an intro about FEP and mention DEM and other methods \cite{friston2008variational}.

%Briefly write about the state estimation literature

%\textcolor{red}{First paragraph could highlight the motivation (the why part) in light of our work's applications and the consequent socio-economic impact. Delivery drones in wind for example.  }, 

% In the field of robotics the popularity of unmanned aerial vehicles (UAV) has been steadily increasing for years. Because of their relatively low cost and high maneuverability, drones are used in a wide variety of applications ranging from military operations to package delivery services and recreational purposes. The ever growing popularity of UAV's has lead to an increase in research into drone control theory, a vital part of which is state observation. With drones often operating outdoors, the influence of wind poses a significant obstacle for steady drone operation and accurate state estimation. Moreover, input prediction under wind conditions provides solutions for fault detection and an approach to handling modeling errors. 

The widespread use of unmanned aerial vehicles (UAV) as delivery drones has increased the need for robust state and input estimators, mainly owing to its safety during uncertain events such as strong wind. We take a step in this direction by evaluating the usefulness of an approach from neuroscience to handle the wind during estimation. 

\begin{figure}[t]
\centering
%\captionsetup{justification=justified}
\includegraphics[scale = 0.4]{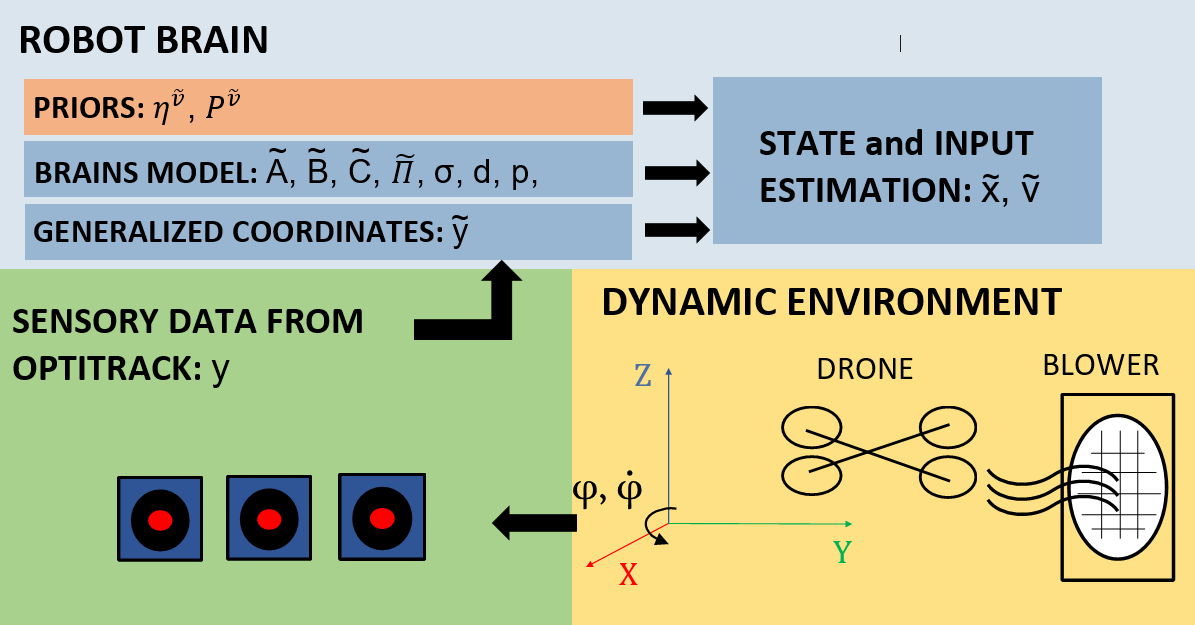}
\caption{The schematic representation of our experimental setup for the DEM's state and input estimation using a quadrotor.}
\label{fig:drone schematic}
\end{figure}

In literature, a wide range of approaches have been used for the state estimation of linear time invariant (LTI) systems. However, most of them assume the noise to be white \cite{white_noise_standard}, which is often a wrong assumption in practice  \cite{brysonSA}. For example, Kalman filter (KF) \cite{gibbs_book} ensures optimality when the noises are white \cite{Kalman}, but it is suboptimal when the noises are colored. An interesting approach from neuroscience called the Free Energy Principle (FEP) uses the concept called generalized coordinates that can leverage the noise derivative information in the brain signals for perception. The FEP based perception scheme called Dynamic Expectation Maximization (DEM) \cite{friston2008variational} was recently reformulated into a state and input observer for LTI systems with colored noise, and was shown to outperform the KF in simulation \cite{meera2020free}. In this paper, we aim to provide the experimental validation of the DEM observer for  a quadrotor under wind conditions using the setup given in Figure \ref{fig:drone schematic}. The main contributions of the paper are:
\begin{enumerate}
    \item Introduce an experimental design with real robots to provide the proof of concept for DEM as a state and input observer.
    \item Provide the first experimental confirmation for the advantage of generalized coordinates in handling colored noise during state and input estimation on robots.
    \item Demonstrate the influence of prior beliefs in shaping the accuracy-complexity trade-off during estimation.
\end{enumerate}

% In literature, a wide variety of solutions for state and input estimation have been presented. However, the standard method for state observation is to model system disturbances as white noise \cite{white_noise_standard}, with one of the most prominent solutions being the Kalman filter \cite{gibbs_book}. The Kalman filter ensures optimality when the white noise assumption is met \cite{Kalman}. However, many real-life applications experience time-correlated, coloured noise \cite{brysonSA}. The coloured noise is often generated by unmodelled system dynamics or external dynamical systems, such as the influence of wind on a drone. The violation of the white noise assumption affects the Kalman filter's optimality and requires alternative solutions for coloured noise handling. Moreover, while solutions for simultaneous state and input estimation do exist, these do not consider coloured noise \cite{state_and_input}. 

% The dynamic expectation maximisation (DEM) method, derived from the free energy principle (FEP), provides a solution for simultaneous state and input estimation in systems under the influence of coloured noise. One of the main strengths of DEM in state and input estimation, comes from its use of generalized coordinates, wherein not only the states but also their derivative information are considered. This enables DEM to capture the time-correlation in coloured noise and to exploit this information for more accurate state and input estimation. This research focuses on the state and input estimation capabilities of DEM on a quadrotor flying under wind conditions. 

%% file: sections/Related_work.tex
\section{RELATED WORK} \label{S:related_work}
%Full literature review of state estimation techniques will be summarized here. Note: no simulation results should appear here. Text only.\\

This section introduces the interdisciplinary nature of FEP, connecting neuroscience and robotics.  

\subsection{Neuroscience} \label{S:related_work_FEP}
%\textcolor{red}{Friston literature}\\
FEP emerges from neuroscience as a unified theory of the brain which posits that all biological systems resist their natural tendency to disorder by minimizing their free energy \cite{friston2010unified}, where free energy is an information theoretic measure that bounds sensory surprisal. FEP provides a mathematical formalism for the brain related functions \cite{friston2009free}, unifies action and perception \cite{friston2010action}, connects memory and attention \cite{friston2009free} and explains Freudian ideas \cite{carhart2010default}. The brain inspired nature of FEP has already attracted roboticists to apply it to build intelligent agents. A few of them includes the body perception of humanoid robots \cite{oliver2019active}, control of manipulator robot \cite{pezzato2020novel}, system identification of a quadrotor \cite{ajith2021DEM_drone,ajith2021convergence}, SLAM \cite{ccatal2021robot} etc. With this work we aim to assess the performance of DEM for state and input estimation of a quadrotor under wind conditions.

\subsection{Robotics and control systems}
%\textcolor{red}{Maybe check for papers with drones flying under wind and see how they deal with it? There should be some state estimators attached to such robust controller designs.}

In control systems literature, numerous approaches are used to deal with colored noise during state estimation. State Augmentation (SA) assumes the colored process noise as an auto-regressive (AR) noise and augments the state space equation to transform it into an equivalent system influenced by white noise  \cite{brysonSA}. The Measurement Differencing \cite{brysonMD} approach deals with handling colored measurement noise. SMIKF \cite{zhouSMIKF} extends KF for coloured noise by incorporating the temporal correlations of the AR noise into the prior covariance calculation of KF. In the fault detection literature, many observers have been developed for input estimation, like the Unknown Input Observer (UIO) \cite{charandabi2012observer}. However, none of these methods perform simultaneous state and input estimation under colored noise other than DEM \cite{meera2020free}.

In robotics, different approaches are used for state estimation of quadrotors under wind conditions. The most common approach (Dryden wind model) is to treat wind as a colored noise shaped by a filter acting on the white noise. Another approach is to model the wind dynamics and estimate wind velocity using complex nonlinear models \cite{drone_wind_wasland}. Using additional cameras for accurate state estimation is another solution \cite{drone_wind_camera}. Our approach differs from these methods as we treat the wind noises as colored and use the information in the noise derivatives for accurate state and input estimation. 

%% file: sections/Problem_statement.tex
\section{PROBLEM STATEMENT} \label{S:problem_statement}
Consider the plant dynamics given in Equation \ref{eqn:general_LTI}, where $A$, $B$ and $C$ are constant system matrices, $\textbf{x}\in \mathbb{R}^n$ is the hidden state, $\textbf{v}\in \mathbb{R}^r$ is the input and $\textbf{y}\in \mathbb{R}^m$ is the output.
\begin{equation} 
\label{eqn:general_LTI}
    \begin{split}
       \Dot{\textbf{x}} = A\textbf{x}+B\textbf{v} + \textbf{w},
    \end{split}
    \quad \quad
    \begin{split}
        \textbf{y} = C\textbf{x} +\textbf{z}.
    \end{split}{}
\end{equation}{}Here $\textbf{w}\in \mathbb{R}^n$ and $\textbf{z}\in \mathbb{R}^m$ are temporally correlated (colored) and represent the process and measurement noise respectively. The noises are assumed to be the result of the convolution of a Gaussian kernel on a white noise signal. The goal of the DEM observer is to simultaneously estimate $\textbf{x}$ and $\textbf{v}$, when the noises are colored (or non-white). The goal of this paper is to design an experimental setup for a real robot that can be used to validate the DEM observer and its usefulness in the presence of colored noise.

% The wind is considered process noise and modelled in $\textbf{w}$ and is much larger in magnitude than $\textbf{z}$. The goal is to estimate $\textbf{x}$ and \textbf{v} as accurately as possible. Before state estimation the process and measurement noise precisions, $\Pi_w$ and $\Pi_z$, are calculated and supplied to the observers. 

%The goal of this paper is to estimate $\textbf{x}$ and \textbf{v} of a quadcopter flying in wind with coloured process noise, given the output $\textbf{y}$, the quadrotor's system matrices $A, B, C$ and the state and output noise precisions $\Pi_w$ and $\Pi_z$. 

%% file: sections/Preliminaries.tex
\section{PRELIMINARIES} \label{S:preliminaries}
%DEM theory goes here. We will detail all the necessary math here.
This section introduces the DEM observer fundamentals.

\subsection{Free energy principle}\label{S:preliminairy_FEP}
% The perception and learning aspects of the FEP are described in DEM \cite{friston2008variational}. In DEM, the brain compares its internal generative model of the environment $m$, described by its internal model parameters $\vartheta$, with the sensory data perceived from the environment $y$. The log-evidence for this model can be expressed as:  
Fundamentally based on Bayesian Inference, FEP estimates the posterior probability $p(\vartheta/y) = {p(\vartheta,y)}/{\int p(\vartheta,y)d\vartheta}$, where $\vartheta$ is the component to be estimated and $y$ is the measurement \cite{buckley2017free}. The presence of an intractable integral motivates the use of a variational density $q(\vartheta)$, called the recognition density that approximates the posterior as $q(\vartheta)\approx p(\vartheta/y)$. This approximation is achieved by minimizing the Kullback-Leibler (KL) divergence of the distributions given by $KL(q(\vartheta)||p(\vartheta/y)) = \langle \ln q(\vartheta)\rangle_{q(\vartheta)} - \langle \ln{p(\vartheta/y)}\rangle_{q(\vartheta)}$, where $\langle.\rangle_{q(\vartheta)}$ represents the expectation over $q(\vartheta)$. Upon simplification using $p(\vartheta/y) = {p(\vartheta,y)}/{p(y)}$, it can be rewritten as \cite{friston2010unified}: 
\begin{equation}
    \label{eqs:logevidence}
    \ln p(y) = F + D_{kl}(q(\vartheta)||p(\vartheta|y)),
\end{equation}
where $ F =  \langle \ln{p(\vartheta,y)}\rangle_{q(\vartheta)}  -\langle \ln q(\vartheta)\rangle_{q(\vartheta)}$ is the free energy. Since $\ln p(y)$ is independent of $\vartheta$, minimization of the KL divergence involves the maximization of free energy. This is the fundamental idea behind using free energy as the proxy for brain's inference through the minimization of its sensory surprisal \cite{friston2010unified}. DEM uses this mathematical framework, in conjunction with the use of generalized coordinates to provide a hierarchical brain model \cite{friston2008hierarchical}. We will be using a reformulated version of DEM given in \cite{meera2020free} for this work.

\subsection{Generative model}
The key concept that differentiates DEM from other methods is its use of generalized coordinates for noise color handling. This is done by keeping track of the trajectory of all time-varying quantities (instead of only its point estimates) through a vector of derivatives. The state vector in generalized coordinates are written using a tilde operator as $\tilde{x}  = [x \ x' \ x'' \ ....]^T$ where the dash operator represents the derivatives. Since the noises are colored, the higher derivatives of the system model can be written as \cite{friston2008variational}: 
% One of the strengths of DEM, its capability to deal with colored noise, comes from the use of generalized coordinates in the generative model. With generalized coordinates, not only the hidden states but also their derivatives are modelled. The result is that the derivatives of the noise are considered, capturing temporal correlation in the noise. The higher order derivatives are modeled as follows:
\begin{equation} 
    \begin{split}
        &x'=Ax+Bv+w \\
        &x''=Ax'+Bv'+w'\\ &...
    \end{split}
    \quad \quad
     \begin{split}
        &y=Cx+z \\
        & y' = Cx'+z'\\ &...
    \end{split}   
\end{equation}{} 
which can be compactly written as: \begin{equation}
\label{eqn:generative_process}
    \begin{split}
        &\dot{\tilde{{x}}}  = D^x\tilde{{x}}= \tilde{A}\tilde{{x}}+\tilde{B}\tilde{{v}} +\tilde{{w}} 
    \end{split}{}
    \quad \quad
    \begin{split}
        &\tilde{{y}} = \tilde{C}\tilde{{x}}+\tilde{{z}}
    \end{split}{}
\end{equation}{} where $D^x = \Bigg[ \begin{smallmatrix}{}
0 & 1 & & &\\
 & 0 & 1 & & \\
 & & .& . &  \\
 & & & 0& 1 \\
 & & & & 0
\end{smallmatrix} \Bigg]_{(p+1)\times (p+1)} \otimes I_{n\times n}.$\\ 
Here, $D^x$ represents the shift matrix, which performs the derivative operation on the generalized state vector. Similarly, $D^v$ performs the same operation on inputs and has size $r(d+1)\times r(d+1)$. $p$ and $d$ represent the embedding order for the hidden states and the inputs respectively, indicating the number of derivatives used. The generalized system matrices are given by $ \tilde A = I_{p+1} \otimes A,\hspace{10pt} \tilde B = I_{p+1} \otimes B,\hspace{10pt}\tilde C = I_{p+1} \otimes C$, where $I$ denotes the identity matrix and $\otimes$ the Kronecker tensor product. The generalized output $\tilde y$ is calculated from the discrete measurements $\hat y = \Bigg[ \begin{smallmatrix} \hdots \\ y(t-dt) \\ y(t) \\ y(t+dt) \\ \hdots \end{smallmatrix} \Bigg]_{m(p+1)}$ using the methodology in \cite{friston2008variational}, resulting in a latency of $\frac{p}{2}dt$ during online estimation, which is negligible for the large sampling rate (120Hz) used in this paper.
\subsection{Noise modeling}
 The use of generalized coordinates helps to model the noise color through the temporal precision matrix of the noise derivatives. In DEM, the noise is assumed to be the result of a white noise signal that has been convoluted using a Gaussian filter of the form: $K(t) = \frac{1}{\sqrt{2\pi}\sigma}exp({-\frac{1}{2}(\frac{t}{\sigma})^2})$. This provides an easy computation of the covariance of the noise derivatives using the temporal precision matrix $S$ \cite{friston2008variational}:
\begin{equation}
\label{eqn:Smatrix}
    S(\sigma^2) = \begin{bmatrix}
    1 & 0 & -\frac{1}{2\sigma^2} & .. \\
    0 & \frac{1}{2\sigma^2} & 0 & ..\\
    -\frac{1}{2\sigma^2} & 0 & \frac{3}{4\sigma^4}& ..\\
    .. & .. & .. & ..
    \end{bmatrix}^{-1}_{(p+1)\times (p+1)}
\end{equation}
$\sigma$ is close to zero for white noise, while $\sigma>0$ for colored noise. The generalized noise precision matrix can be written using $S$ as $\tilde{\Pi} = diag(\tilde \Pi^z ,P^{\tilde{v}},\tilde \Pi^w )$, where $\tilde \Pi^z = S \otimes \Pi^z$, $\tilde \Pi^w = S \otimes \Pi^w$, and $P^{\tilde v} = S \otimes P^v$. Here $\Pi^w$ and $\Pi^z$ are the noise precisions (inverse covariance), and $P^v$ is the prior precision on inputs. 

% \subsection{Generalized output}
% The generalized output $\tilde y$ of equation \ref{eqn:generative_process} is often unavailable and expensive in real life applications. This problem is solved by computing $\tilde{y}$ from the discrete output measurements $\hat{y}$ using the Taylor expansion (and then inverting it):
% \begin{equation}
% \label{eqn:y_taylor}
%     \hat y = \begin{bmatrix} \hdots \\ y(t-dt) \\ y(t) \\ y(t+dt) \\ \hdots \end{bmatrix}_{m(p+1)} = (E \otimes I_m)\tilde y,
% \end{equation}
% where $E$ is calculated as:
% \begin{equation}
%     \label{eqn:eqtaylor}
%     E_{ij} = \frac{[(i-\frac{p+1}{2})dt]^{j-1}}{(j-1)!}.
% \end{equation}
% Here, $i$ and $j$ are integers from 1 to $p + 1$. Rewriting Equation \ref{eqn:y_taylor} yields:
% \begin{equation}
% \label{eq:rev_taylor}
%     \tilde y = (E^{-1}\otimes I_m)\hat y
% \end{equation}
% The use of discrete measurements to compute $\tilde{y}$ comes with a latency of $p$ in state estimation. 
%With the definitions from sections A to D, it is possible to formulate the DEM state and input observer equations.

\subsection{State and Input Observer}
The DEM observer in \cite{meera2020free} simultaneously estimates the generalized state and input vector $X = \big[ \begin{smallmatrix} \tilde{x} \\ \tilde{v} \end{smallmatrix}$\big] through the gradient ascend over its variational free energy $V(t)$:
\begin{equation}
\label{eqn:gradient_ascent}
    \dot X = k V(t)_X + DX,
\end{equation}
where $k$ is the learning rate, $V(t)_X$ is the gradient of $V(t)$ with respect to $X$ and $D^X = \begin{bmatrix} D^x & O \\ O & D^v \end{bmatrix}$. Using the Laplace approximation \cite{friston2007variational}, simplifies $V(t)$ as the precision weighted prediction error,
$    V(t) = -\frac{1}{2}\tilde \epsilon^T\tilde \Pi \tilde \epsilon$, where $\tilde \epsilon$ is the prediction error given by:
\begin{equation}
    \tilde \epsilon = \begin{bmatrix}\tilde y - \tilde C \tilde x \\ \tilde v -  \eta^{\tilde{v}} \\ D^x\tilde x - \tilde A \tilde x - \tilde B \tilde v \end{bmatrix}
\end{equation}
Here $\eta^{\tilde{v}}$ denotes the prior on the input. Therefore, $V(t)_X = -\tilde \epsilon_X^T\tilde \Pi \tilde \epsilon$, where $\tilde \epsilon_X$ is given by:
\begin{equation}
    \tilde \epsilon_X = \begin{bmatrix} -\tilde C & O\\ O & I\\ D^x-\tilde A & - \tilde B\end{bmatrix}.
\end{equation}
%Laplace approximation for the conditional density:
%\begin{equation}
%    U(y,\vartheta) \approx U(\mu,y) + (\vartheta - \mu)^TU_{\vartheta\vartheta}(\vartheta - \mu)
%\end{equation}
Substituting these results to Equation \ref{eqn:gradient_ascent} upon simplification yields the DEM state and input observer of \cite{meera2020free}:
\begin{equation}  \label{eqn:D_linear_observer}
\dot{X} = \begin{bmatrix}
\dot{\tilde{x}} \\ \dot{\tilde{v}} 
\end{bmatrix} = A_1 \begin{bmatrix}
\tilde{x}\\ \tilde{v}
\end{bmatrix} + B_1 \begin{bmatrix}
\tilde{\textbf{y}} \\ -\tilde{\eta}
\end{bmatrix} and \ Y=X,
\end{equation}
where $Y$ is the output of the observer, $A_1 = D^X - k A_2$,  
\begin{equation} \label{eqn:A2}
     A_2 = \begin{bmatrix}
                \tilde{C}^T\tilde{\Pi}^z\tilde{C}+(D^A )^T\tilde{\Pi}^w D^A  & -(D^A )^T\tilde{\Pi}^w\tilde{B} \\
                -\tilde{B}^T \tilde{\Pi}^w D^A  & \tilde{P}^v+\tilde{B}^T \tilde{\Pi}^w \tilde{B}
            \end{bmatrix},
\end{equation}
$B_1  = -\begin{bmatrix}
                -\tilde{C}^T\tilde{\Pi}^z & O \\ O & \tilde{P}^v
            \end{bmatrix}$ and $D^A = D^x - \tilde{A}$.

This observer was proved to outperform the KF for state estimation on LTI systems with colored noise in simulation \cite{meera2020free}. We will use this observer design throughout the paper to provide the experimental validation on a real robot. 
% Equation \ref{eqn:D_linear_observer} is the DEM state and input observer. For state estimation in the discrete domain, the system from equation \ref{eqn:D_linear_observer} is discretized. This observer was already tested on simulated systems \cite{meera2020free}, where it was compared to the Kalman filter. In the next section, the experimental setup of the current research will be elaborated. 

\subsection{Uncertainty in state and input estimation}
DEM provides a means to compute the uncertainty in estimation through the precision of estimates given by the negative curvature of variational free energy \cite{anilmeera2021DEM_LTI}:
\begin{equation}
    \Pi^X = -V(t)_{XX} = \tilde \epsilon_X^T\tilde \Pi \tilde \epsilon_X = A_2,
\end{equation}
where $A_2$ is given in Equation \ref{eqn:A2}. Therefore, the precision of DEM's state and input estimates is independent of time, and is given by $\Pi^{\tilde{x}\tilde{x}} =  \tilde{C}^T\tilde{\Pi}^z\tilde{C}+(D^A )^T\tilde{\Pi}^w D^A $ and $\Pi^{\tilde{v}\tilde{v}} = \tilde{P}^v+\tilde{B}^T \tilde{\Pi}^w \tilde{B}$ respectively.

%% file: sections/Experimental_design.tex
% \begin{figure}[htpb]
% \centering
% \captionsetup{justification=justified}
% \includegraphics[scale = 0.5]{pics/drone.jpg}
% \caption{The AR drone 2.0 used for the experiments.}
% \label{fig:drone}
% \end{figure}

\section{EXPERIMENTAL DESIGN}
The distinctive feature of DEM that enables it to handle colored noise (to outperform a KF for state estimation) is its use of generalized coordinates \cite{friston2008variational,meera2020free}. This section aims at designing an experimental setup (as simple as possible) for real robots that can leverage this property and provide a proof of concept for our DEM-based state and input observer design for LTI systems with colored noise \cite{meera2020free}.

\subsection{Experimental setup}
\begin{figure}[htpb]
\centering
\captionsetup{justification=justified}
\includegraphics[scale = 0.4]{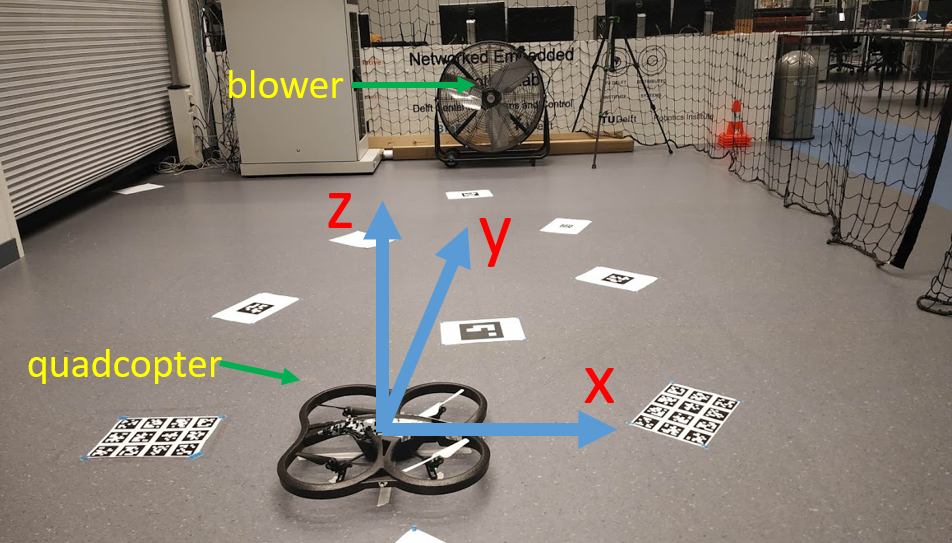}
\caption{The controlled lab environment for the experimental setup with the quadrotor and the blower.}
\label{fig:lab}
\end{figure}

Our experimental setup consist of a quadrotor (Parrot AR.drone 2.0) hovering in wind produced by a blower in a controlled lab, as shown in Figure \ref{fig:lab}. The blower induces wind in the negative $y$ direction, against the hovering quadcopter. The PID controller tries to resist the wind to hover the quadrotor at the given position $(0m,0m,1m)$ and orientation $(0^{\circ},0^{\circ},0^{\circ})$. We use an OptiTrack motion capture system to record the position and orientation of the quadcopter. A total of 9 hovering experiments were performed - 4 experiments without wind (blower off) and 4 experiments with wind (blower on). The final experiment was used to tune all the benchmark observers and will not be used for benchmarking.  Each experiment lasted $10s$ with $dt=0.0083s$.

Since wind is the result of another (unmodelled) dynamic system, we hypothesize that the introduction of wind influences the quadrotor dynamics and acts as the source of colored noise to the system. The experimental design enables us to control the level of noise color entering the system by controlling the blower for its wind speed and direction. We hypothesize that our DEM observer can leverage on the information contained in the colored noise by keeping track of the higher derivatives of states and inputs through the generalized coordinates.

% Big drone or small drone state space? 
% How much detail on the experiment? Also on what data was chosen to analyze, or just analyze everything? 

% The goal of the experiment is to obtain data from a quadrotor drone, which could be analyzed and used for state and input estimation. The drone that was used for these experiments is a Parrot AR.drone 2.0, shown in figure \ref{fig:drone}.

% The drone was tasked to keep a constant position and orientation while hovering in the air. After flying into position, a blower was used to supply wind to the drone, providing a large disturbance to the system. A picture of the experimental setup, including the drone and the blower, is shown in figure \ref{fig:lab}. The drone's position and orientation were externally monitored using an OptiTrack motion caption system. The data was collected via the Robot Operating system (ROS), and stored into ROS bag-files, which were processed and examined in Matlab.

\begin{figure*}[t!]
    \centering
    \begin{subfigure}[b]{0.32\textwidth}
		\includegraphics[width=\textwidth,trim={0cm 0cm 0cm 0},clip]{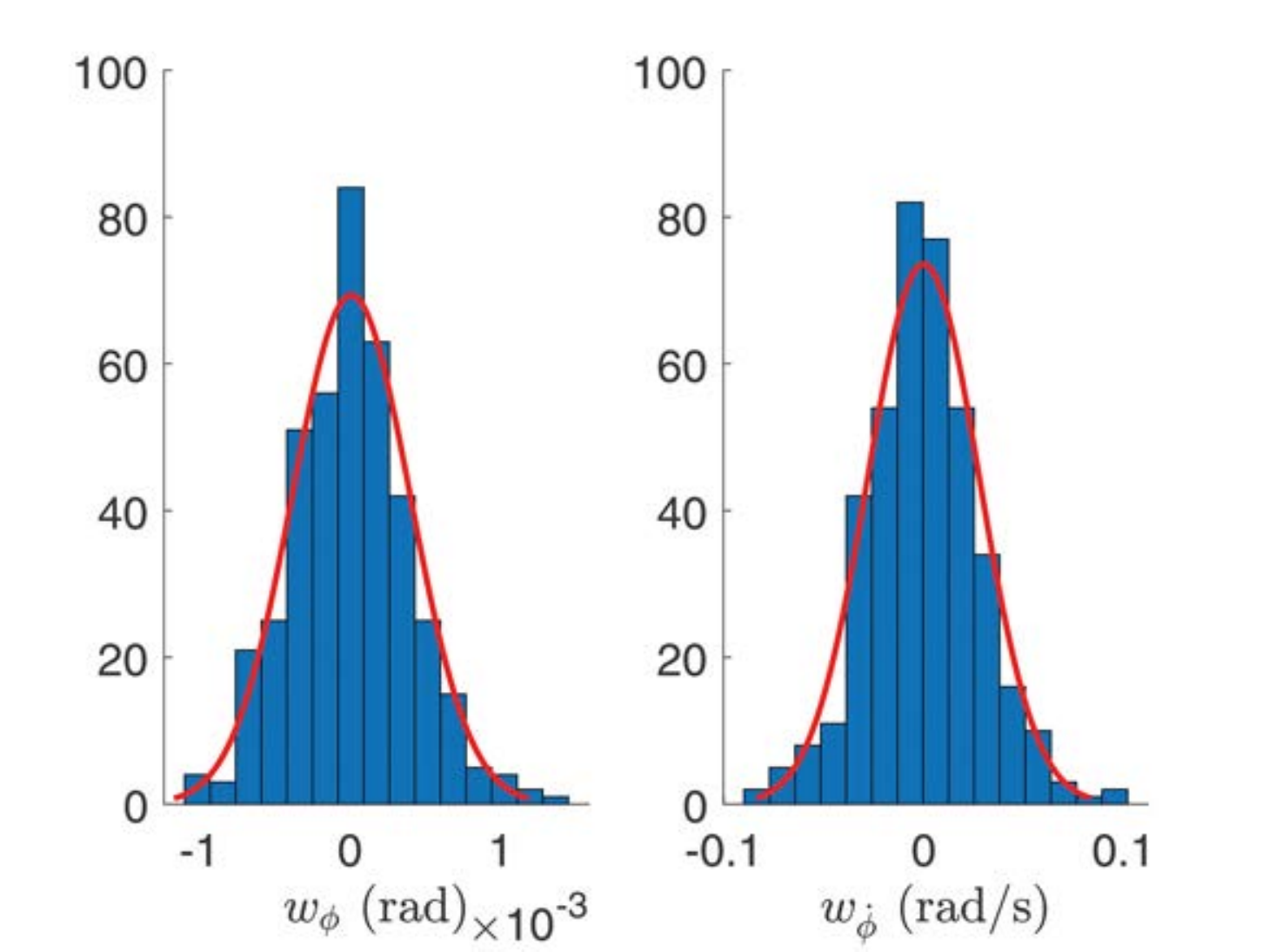}
        \caption{Histograms of the process noise $w_{\phi}$ and $w_{\dot \phi}$ with a Gaussian fit for no wind conditions.}\label{fig:hist_NW}
    \end{subfigure}    
   \begin{subfigure}[b]{0.32\textwidth}
		\includegraphics[width=\textwidth,trim={0cm 0 0cm 0},clip]{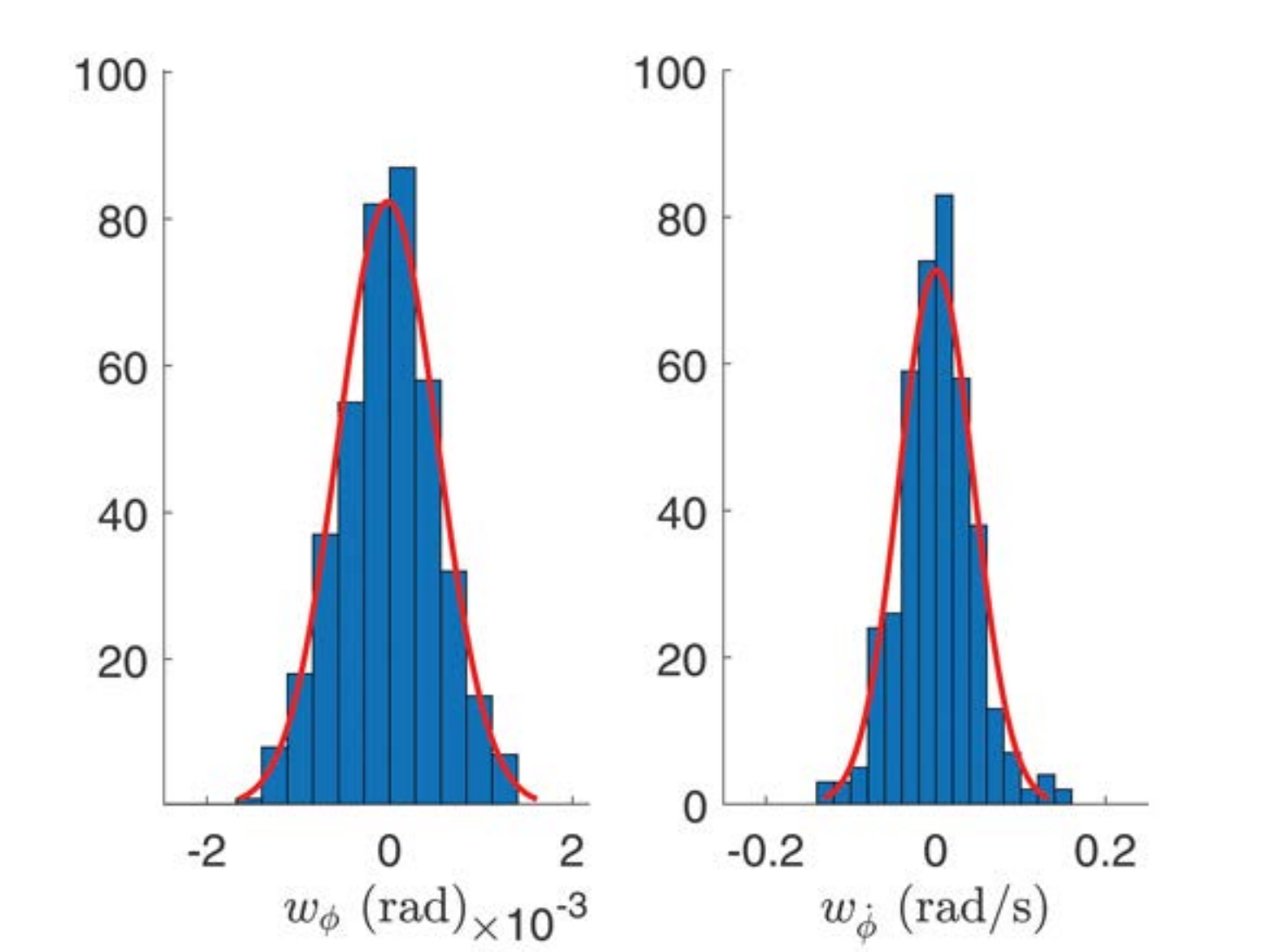}
        \caption{Histograms of the process noise $w_{\phi}$ and $w_{\dot \phi}$ with a Gaussian fit under wind conditions.}\label{fig:hist_WW}
   \end{subfigure}   
       \begin{subfigure}[b]{0.32\textwidth}
		\includegraphics[width=\textwidth,trim={0cm 0cm 0cm 0cm},clip]{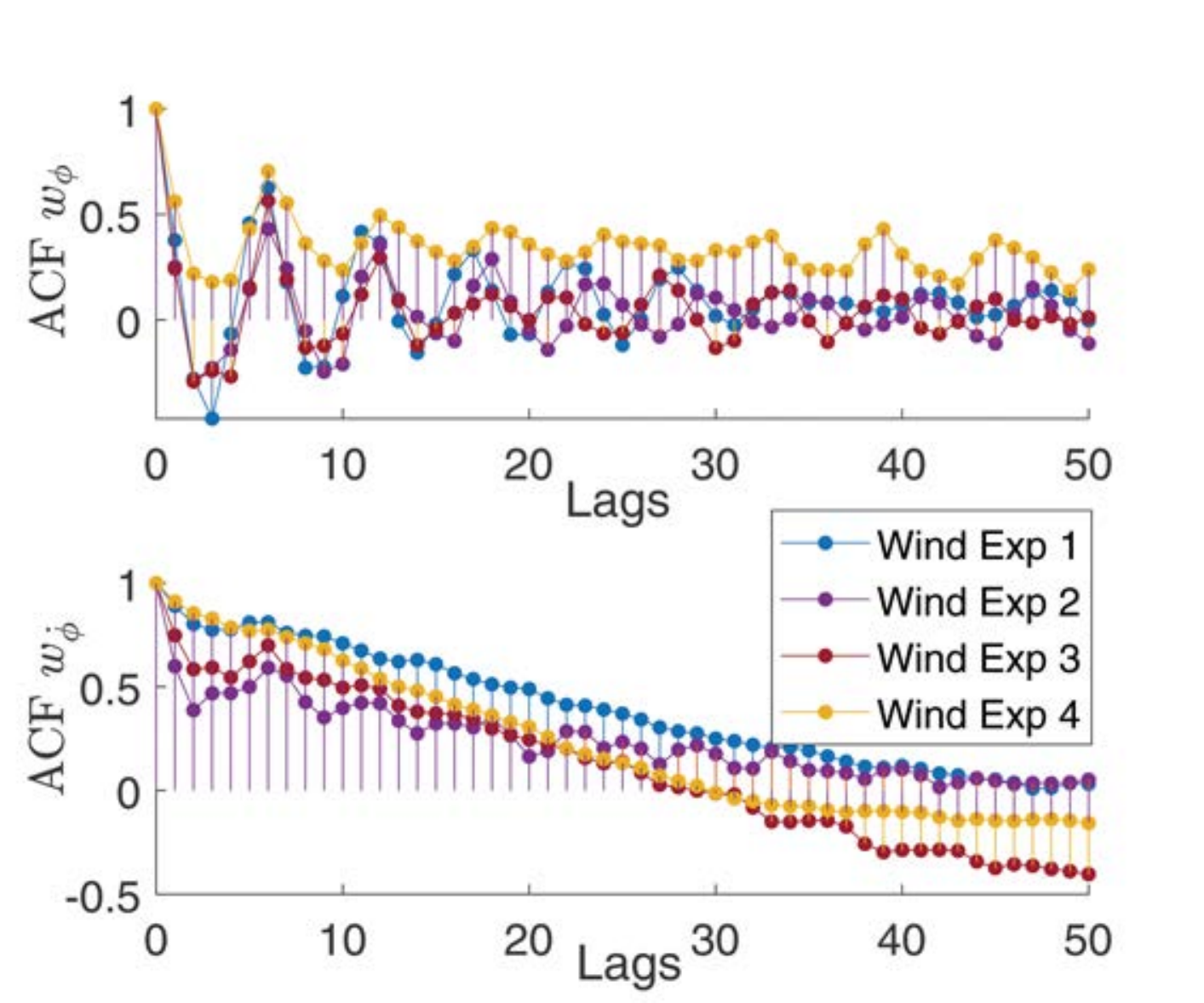}
        \caption{The auto-correlations for the process noises $w_{\phi}$ and $w_{\dot \phi}$ under wind conditions.}\label{fig:auto_corr}
    \end{subfigure} 
    \caption{The properties of process noise of our experiment. The wind introduces a colored Gaussian distributed disturbance to the system.}\label{fig:Noise_charact}
\end{figure*}

\subsection{Quadrotor model selection}
The quadrotor model selection was performed to accommodate the influence of wind using minimum number of states, resulting in a controllable and observable LTI system. Since the wind flows in negative $y$ direction, it influences the roll angle ($\phi$ around $x$-axis) and the roll angular velocity ($\dot{\phi}$) the most. Therefore, we only consider states $x = \Big[ \begin{smallmatrix}
\phi \\ \dot{\phi}
\end{smallmatrix} \Big] $.
The model involving these states is based on the one provided in \cite{bouabdallah2004design}. By assuming small angles, $\phi$ and $\dot{\phi}$ can be decoupled from the other system dynamics. Linearizing these states around hovering conditions gives:
% Since the roll angle around the x-axis is perpendicular to the wind direction, this angle experiences the largest wind disturbance. Therefore the system will be limited to the roll angle and roll rate, $\phi$ and $\dot \phi$. A linearized state-space model for the drone's roll angle and rate is given in equation \ref{eqn:ss_small}.
% The quadrotor model can be written as:
% \begin{equation*}
%     \textcolor{red}{add \ non-linear \ model \ here}
% \end{equation*}
% which can be linearized around the equilibrium point as:
\begin{equation}
\label{eqn:ss_small}
\begin{aligned}
    \begin{bmatrix}
    \dot \phi\\
    \ddot \phi
    \end{bmatrix} &= \begin{bmatrix}
    0&1\\0&0
    \end{bmatrix}\begin{bmatrix}
    \phi \\ \dot \phi
    \end{bmatrix} + \begin{bmatrix}
        0&0&0&0\\
        \frac{c_{B\phi}}{I_{xx}}&-\frac{c_{B\phi}}{I_{xx}}&-\frac{c_{B\phi}}{I_{xx}}&\frac{c_{B\phi}}{I_{xx}}
    \end{bmatrix} \Bigg[ \begin{smallmatrix}
    pwm_1\\pwm_2\\pwm_3\\pwm_4
    \end{smallmatrix} \Bigg] \\
    y &= \begin{bmatrix}
    1&0
    \end{bmatrix}\begin{bmatrix}
    \phi\\
    \dot \phi
    \end{bmatrix}
\end{aligned}
\end{equation}
Here $pwm_i$ is the Pulse Width Modulation signal provided to the $i^{\text{th}}$ motor by the controller for stable hovering. $I_{xx}$ is the quadcopter's moment of inertia around the $x$-axis. It's value is identified using the bifilar pendulum experiment and equals $3.4 \cdot 10^{-3} kgm^2$. $c_{B\phi}$ is the thrust coefficient that models the relation between the PWM values and the thrust generated by the quadcopter rotors. It's value is obtained by averaging the results of several static thrust tests and equals $1.274 \cdot 10^{-3}Nm$.
% In this equation, the $pwmA$ values are the motor Pulse Width Modulation (PWM) values that are applied to the drone as input. $I_{xx}$ is the drone's moment of inertia around the X-axis. Finally, $c_{B\phi}$ is a thrust coefficient that models the relation between the PWM values and the thrust, generated by the drone's rotors. In this experiment, the value for $C_{B\phi}/I_{xx}$ is $0.3748$. For the data analysis the state-space system is transformed to the discrete domain in the form:
% \begin{equation}
%     \begin{aligned}
%     x_{k+1} &= A_d x_k + B_d v_k + w_k\\
%     y_k &= C_d x_k + z_k
%     \end{aligned}
% \end{equation}
% In this equation, $A_d$, $B_d$,and $C_d$ are the discretized system matrices, $x$ is the state containing $\phi$ and $\dot \phi$, $y$ is the measurement containing $\phi$ measured from the OptiTrack system, and $v$ represents the inputs, consisting of the applied PWM values, as measured by the drone. 
We normalize the input $pwm$  signals using $ v = \frac{v-mean(v)}{max(v)-min(v)} $ and use the same factor to multiply the $B$ matrix, such that the system dynamics are unaltered. See \cite{benders2020ar} for more details regarding the model derivation and system identification procedure.
% In order to accommodate for more accurate input estimation, the inputs are normalized by dividing them by a normalizing constant and multiplying the B-matrix with said constant. Moreover since the input values contained a non-zero mean, because the rotors have to counter gravity, the mean was subtracted from the inputs. This did not effect the current system, since only the roll angles are examined and not the y position or velocity. 

Since we use an accurate measurement system, $\Pi^z$ is very high for all experiments. However, the presence of colored process noise $w$ through wind makes $ \Pi^w << \Pi^z $. $\Pi^w$ is further influenced by the modelling errors during linearization as the wind aggressively drives the quadrotor away from its equilibrium.

%% file: sections/Results.tex
\section{RESULTS AND ANALYSIS} \label{S:experimental_results}
This section aims to investigate the validity of the assumptions in our experimental design and to compare the performance of DEM observer against other benchmarks.

% This section summarizes the results that were obtained during the experiments and explores the state and input estimation capabilities of DEM. 

%\subsection{Tuning the s value}
%As discussed in section \ref{SS:noise_param} the $s$-value used for the remainder of the experiments is tuned on a separate experiment. DEM state estimation is performed for a variety of $s$-values. The best state estimation results in terms of SSE are obtained for an $s$-value of 0.006. This value is used in the remainder of the experiments to obtain the results. \textcolor{red}{Quite verbose and scientifically uninformative paragraph}

\begin{figure*}[ht!]
    \centering
    \begin{subfigure}[b]{0.32\textwidth}
		\includegraphics[width=\textwidth,trim={0cm 0 0cm 0},clip]{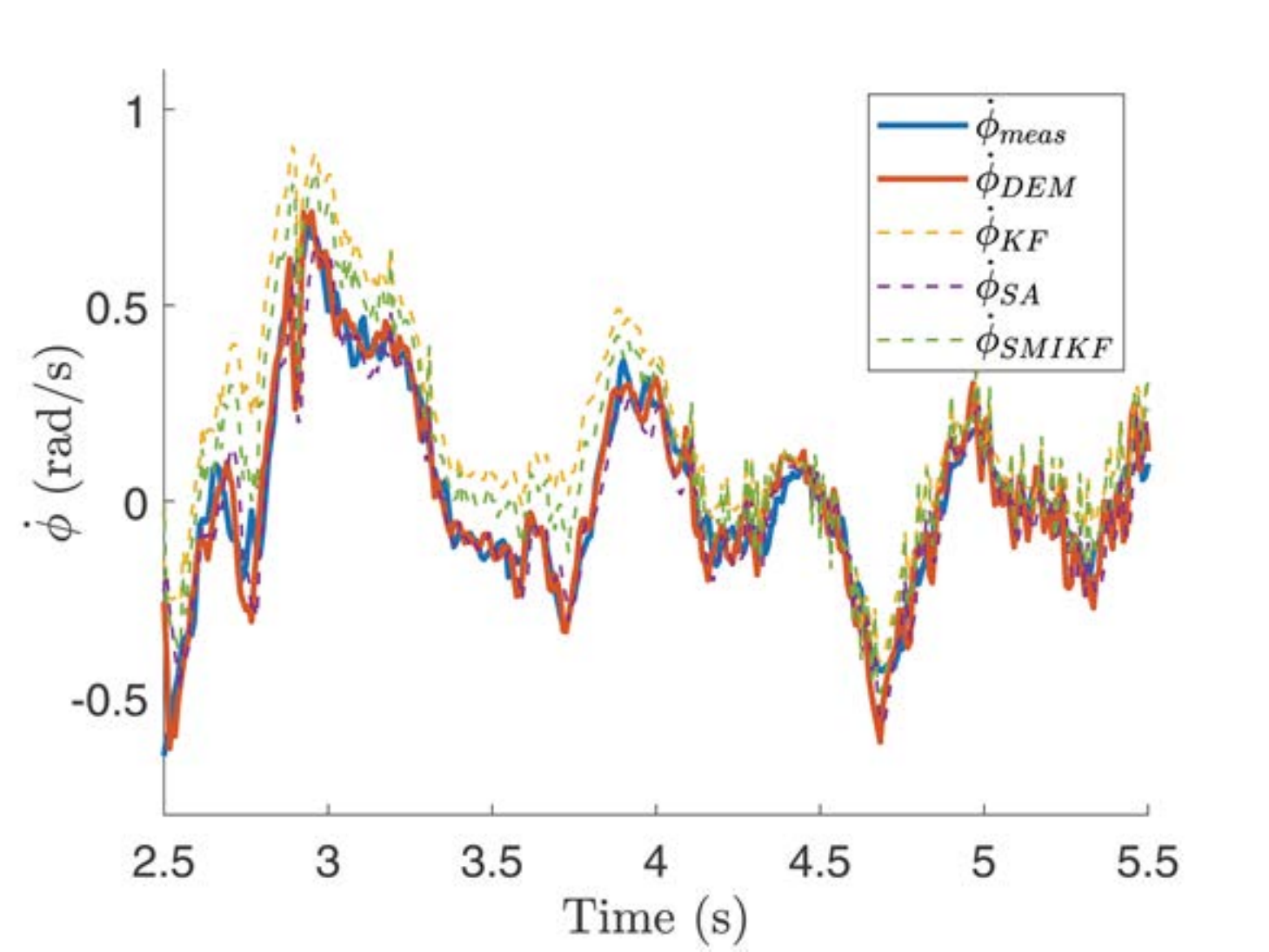}
		\caption{State estimation benchmarks (with wind).}\label{fig:state_est}
   \end{subfigure}   
    \begin{subfigure}[b]{0.3\textwidth}
		\includegraphics[width=\textwidth,trim={0cm 0cm 0cm 0cm},clip]{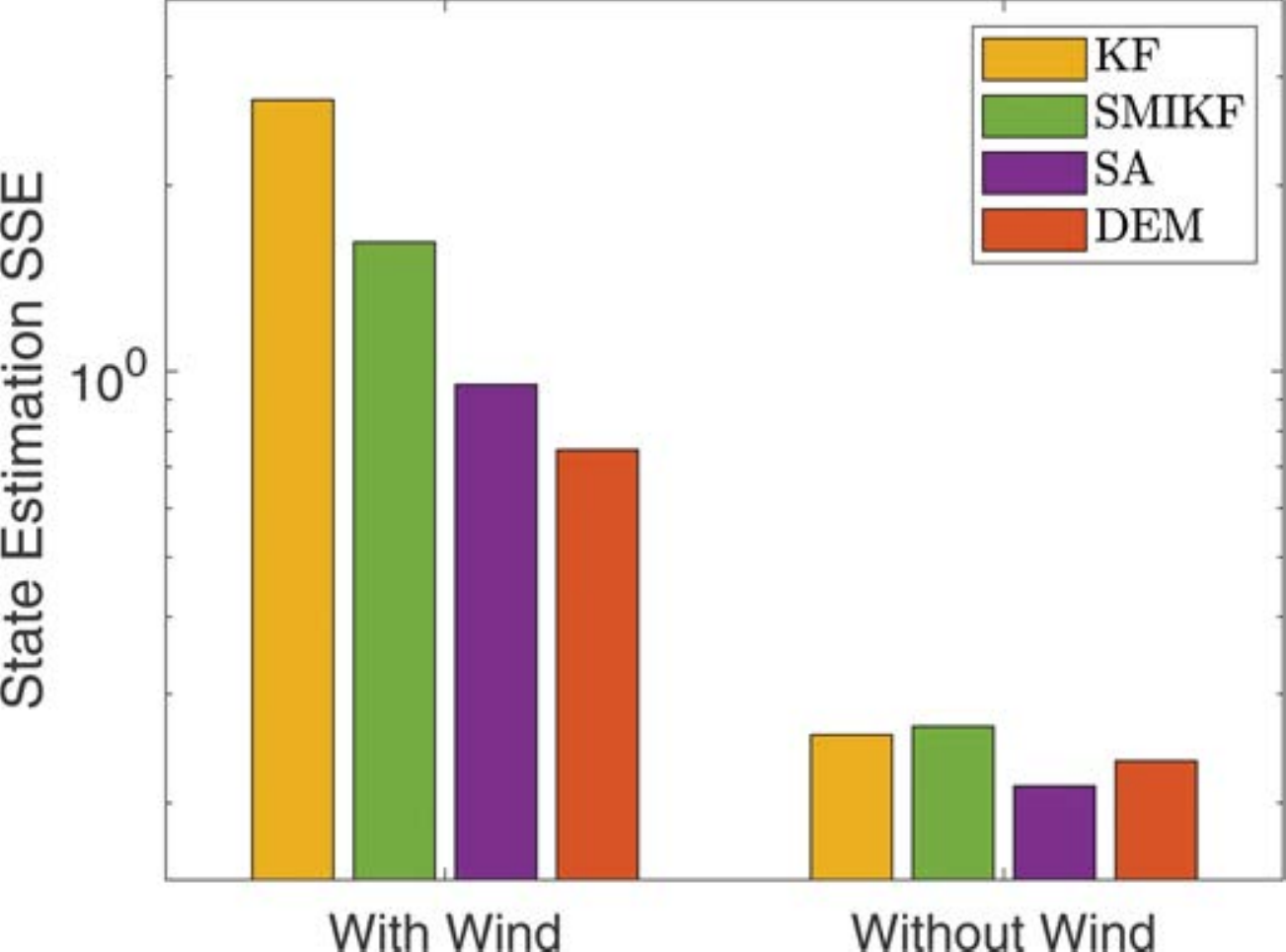}
		\vspace{0pt}
		\caption{Average SSE of all flights.}\label{fig:bench_bar}
    \end{subfigure} 
    \begin{subfigure}[b]{0.31\textwidth}
		\includegraphics[width=\textwidth,trim={0cm 0cm 0cm 0},clip]{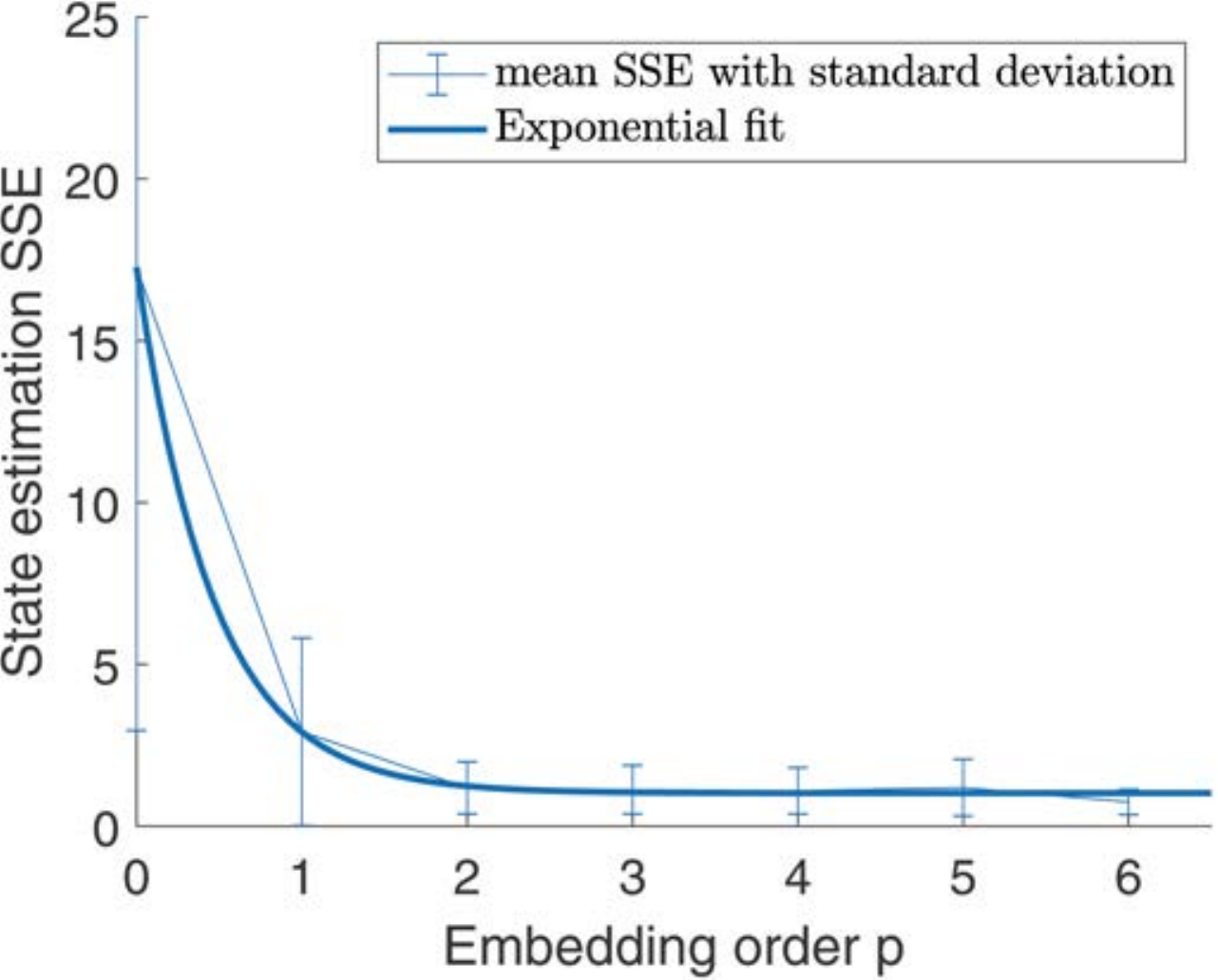}
		\caption{Average SSE drops exponentially with $p$.}\label{fig:vary_p}
    \end{subfigure}   
    \caption{DEM outperforms other benchmarks with minimal SSE in state estimation for a quadrotor flying under windy conditions. The performance of DEM improves exponentially with higher orders of generalized motion $p$, for a quadrotor flying under windy conditions, highlighting the importance of generalized coordinates in the presence of colored noise.  }\label{fig:v_estimate_sigma}
\end{figure*}

\subsection{Validity of Laplace Approximation}
The DEM framework approximates the probability densities of $p(\tilde{y})$ and $p(\tilde{x}/\tilde{v}) $ to be Gaussian in nature, centred around their mean predictions ($\tilde{C}\tilde{x}$ and $ \tilde{A}\tilde{x} + \tilde{B}\tilde{v}$, respectively) with the same precision as that of the noises ($\tilde{\Pi}^z$ and $\tilde{\Pi}^w$):
\begin{equation} \label{eqn:p(y/theta)}
\begin{split}
        p(\tilde{y}) & = \frac{1}{\sqrt{(2\pi)^{m(p+1)}|\tilde{\Sigma}^z|}}e^{-\frac{1}{2}(\tilde{y}-\tilde{C}\tilde{x})^T \tilde{\Pi}^z (\tilde{y}-\tilde{C}\tilde{x})},\\
        p(\tilde{x}/\tilde{v}) & = \frac{1}{\sqrt{(2\pi)^{n(p+1)}|\tilde{\Sigma}^w|}}e^{-\frac{1}{2}\tilde{\epsilon}^{xT} \tilde{\Pi}^w \tilde{\epsilon}^x},
\end{split}
\end{equation}
where $\tilde{\epsilon}^x = D^x\tilde{x} - \tilde{A}\tilde{x} - \tilde{B}\tilde{v}$. The validity of this approximation on our experimental design was investigated by plotting the process noise histograms for both without wind and with wind conditions (for 400 data points each) and is shown in Figure \ref{fig:hist_NW} and \ref{fig:hist_WW} respectively. Similar trend holds for measurement noise as well. The strong Gaussian fit indicates the validity of Laplace approximation for our experimental design.

% As discussed in section \ref{S:preliminaries}, DEM uses the Laplace approximation to leverage the Gaussian nature of Bayesian inference to an LTI control problem. Therefore, an important step in benchmarking DEM on experimental data is to verify that the noises follow a Gaussian distribution. In figures \ref{fig:hist_NW} and \ref{fig:hist_WW} the histograms of the process noises $w_{\phi}$ and $w_{\dot\phi}$ have been presented for the experiments without and the experiments with wind. As can be observed from the figures, the noise follows a Gaussian distribution which accommodates for use of the Laplace assumption in DEM. Moreover, it can be observed that adding wind to the system increases the noise while maintaining the Gaussian distribution. In combination with the results from section \ref{SS:colored_noise} it can be concluded that the system is influenced by primarily Gaussian colored noise. 

\subsection{Influence of wind on states and process noise}
In this section we validate the direct influence of wind on the states and process noise. Table \ref{tab:std} demonstrates a higher standard deviation for windy conditions than for non-windy conditions. A similar trend can be observed from the width of histograms in Figure \ref{fig:hist_NW} and \ref{fig:hist_WW}, indicating that our experimental design can control the noise generation.
\begin{table}[h]
    \centering
    \begin{tabular}{|c|c|c|c|c|c|}
    \hline
    & $\phi$ (rad) & $\dot{\phi}$ (rad/s) & $w_{\phi}$ (rad) & $w_{\dot\phi}$ (rad/s) \\    
    \hline
    Without wind & 0.00855 & 0.0544 & 0.000416 & 0.0284 \\ 
    \hline
     With wind & 0.0460 & 0.260 & 0.000937 & 0.0607  \\
    \hline
    \end{tabular}
    \caption{The standard deviations of the states, $\phi$ and $\dot\phi$, and the process noises, $w_{\phi}$ and $w_{\dot\phi}$, for experiments with and without wind.}
    \label{tab:std}
\end{table}

\subsection{Confirmation of noise color}\label{SS:colored_noise}
In this section we confirm that our experimental design generates colored process noise. Figure \ref{fig:auto_corr} shows the sample auto-correlation of the process noise of all experiments (with wind). There is stronger autocorrelation for $w_{\dot{\phi}}$ than for $w_{\phi}$, because $\phi$ is observed. The auto-correlation is different from that expected from a white noise signal where the auto-correlation immediately drops to 0 after zero lag. This confirms the presence of strong noise color (time-correlated noise) in data. 

% The noise was isolated as discussed in section \ref{SS:noise_param}. The standard deviation of the measurements noise, $\sigma_z$, is $9.02\cdot10^{-5}$. While the standard deviations of the states and process noises are reported in table \ref{tab:std}.

%the noise was obtained by equation \ref{eqn:noise_w}. The standard deviation of the measurement noise was determined while the drone was stationary and on the ground. In the experiments, the standard deviation of the measurement noise, $\sigma_z$, was $9.02\cdot10^{-5}$. In contrast, the standard deviation of the process noise, $w_{\phi}$, was $3.280 \cdot 10^{-4}\hspace{5pt} rad$ without wind and $0.0010 \hspace{5pt}rad$ with wind. The standard deviation of $w_{\dot \phi}$ was $0.0265 \hspace{5pt}rad/s$ without wind and $0.0935\hspace{5pt} rad/s$ with wind. 

% From these results it can be observed that the process noise is much larger in magnitude than the measurement noise and that by introducing wind to the system, a large disturbance is added to the states. For further investigation, the autocorrelation plots of the process noises with wind are given in figure \ref{fig:auto_corr}. From these graphs it can be observed that the process noises in both $\phi$ and $\dot \phi$ are autocorrelated over time, indicating that the noise is colored. 

\subsection{Estimator settings for benchmarking} \label{sec:settings}
We aim to benchmark the state estimation against KF, SMIKF and SA, and the input estimation against UIO, for a total of 8 experiments (4 with and 4 without wind). DEM was set with the order of generalized motion of states and inputs to $p=6$ and  $d=2$ respectively. The SMIKF implementation could only accommodate a first order AR model, while the SA implementation uses a 6$^{th}$ order AR model, similar to the 6$^{th}$ order derivatives ($p$) of DEM. The noise precision $\Pi^w$ was calculated for each experiment, while $\Pi^z = 8.1\cdot10^{-9}$ was calculated from static drone data. The 9$^{th}$ experiment was used to tune the noise smoothness to $\sigma=0.006$, which was used for all experiments. 

%The noise smoothness was tuned to $\sigma=0.006$ using the ninth experiment and was kept the same for all experiments.

% After examining the noise characteristics for wind and no wind conditions and determining the $s$-value as discussed in section \ref{SS:noise_param} the DEM algorithm is tuned as follows:
% \begin{itemize}
%     \item $p = 6$
%     \item $d = 2$
%     \item $\sigma_v = e^{-16}$ (for known input)
%     \item $s = 0.006$
% \end{itemize}
% The noise precisions $\Pi_w$ and $\Pi_z$ are determined per experiment as described in section \ref{SS:noise_param}. The priors on the input, $\tilde \mu$, are set as the embedded measured input $\tilde v$ to accommodate for state estimation with known inputs. With these settings DEM state estimation is performed on all 8 experiments, 4 without wind conditions and 4 w with wind conditions. The experiments are divided into 5 segments of 2 seconds to accommodate for outlier detection. Next to DEM, the Kalman filter, SMIKF and the SA filter are also used to perform state estimation. SMIKF is limited to a first order AR model but SA was set to a sixth order AR model, in order to reflect the level of noise modeling of the sixth order embedded derivatives of DEM. 

\subsection{State estimation - benchmarking}
In this section, we compare the performance of DEM with the aforementioned benchmarks for state estimation with known inputs. Figure \ref{fig:state_est} shows the state estimates of all benchmarks for an experiment with wind. Although most benchmarks follow the general trend of the measured states (in blue), DEM performs the best. KF shows an inferior performance due to its incapability of dealing with colored noise. We use the sum of squared errors (SSE) between the estimate of $\dot \phi$ and its measurement as the metric to denote the quality of state estimation. The average SSE of all 4 experiments (with and without wind separately) for all benchmarks are shown in Figure \ref{fig:bench_bar}. DEM outperforms other benchmarks in state estimation under wind conditions with minimal SSE, demonstrating that it is a competitive state estimator. 

% The methods are benchmarked based on the sum of squared errors (SSE) between the estimate of $\dot \phi$ and the measured state. An example of the state estimations is depicted in figure \ref{fig:state_est}, where the real state, the DEM estimate and the benchmark estimates are plotted together. The resulting mean SSE's of the experiments without and with wind are given in figure \ref{fig:bench_bar}. From these two figures it can be observed that for wind conditions, DEM is able to capture the real state the most accurate, which results in the lowest SSE, thereby outperforming the other benchmarks for wind conditions.  
%(\textcolor{red}{The text explaining the result, isn't sufficient to understand how DEM "outperforms". Be as direct as possible. Also, you haven't really introduced the metric.}). Under no wind conditions, where the noise is less colored, SA still outperforms DEM. 

\subsection{Role of Generalized Coordinates}
One of the main strengths of DEM - the capability to deal with colored noise - comes from the use of generalized coordinates. In this section, we demonstrate the usefulness of generalized coordinates in state estimation on experimental data. The mean (and standard deviations of) SSE of state estimation for  all experiments (with wind)  for varying orders of generalized motion $p$ is shown in Figure \ref{fig:vary_p}. The exponential decrease in SSE is consistent with the results from \cite{meera2020free} on large simulated data, and indicates the importance of generalized coordinates in accurate state estimation in the presence of colored noise. 

\subsection{State estimation as free energy maximization}
\begin{figure}[htb]
    \centering
    \captionsetup{justification=justified}
    \includegraphics[scale = 0.45]{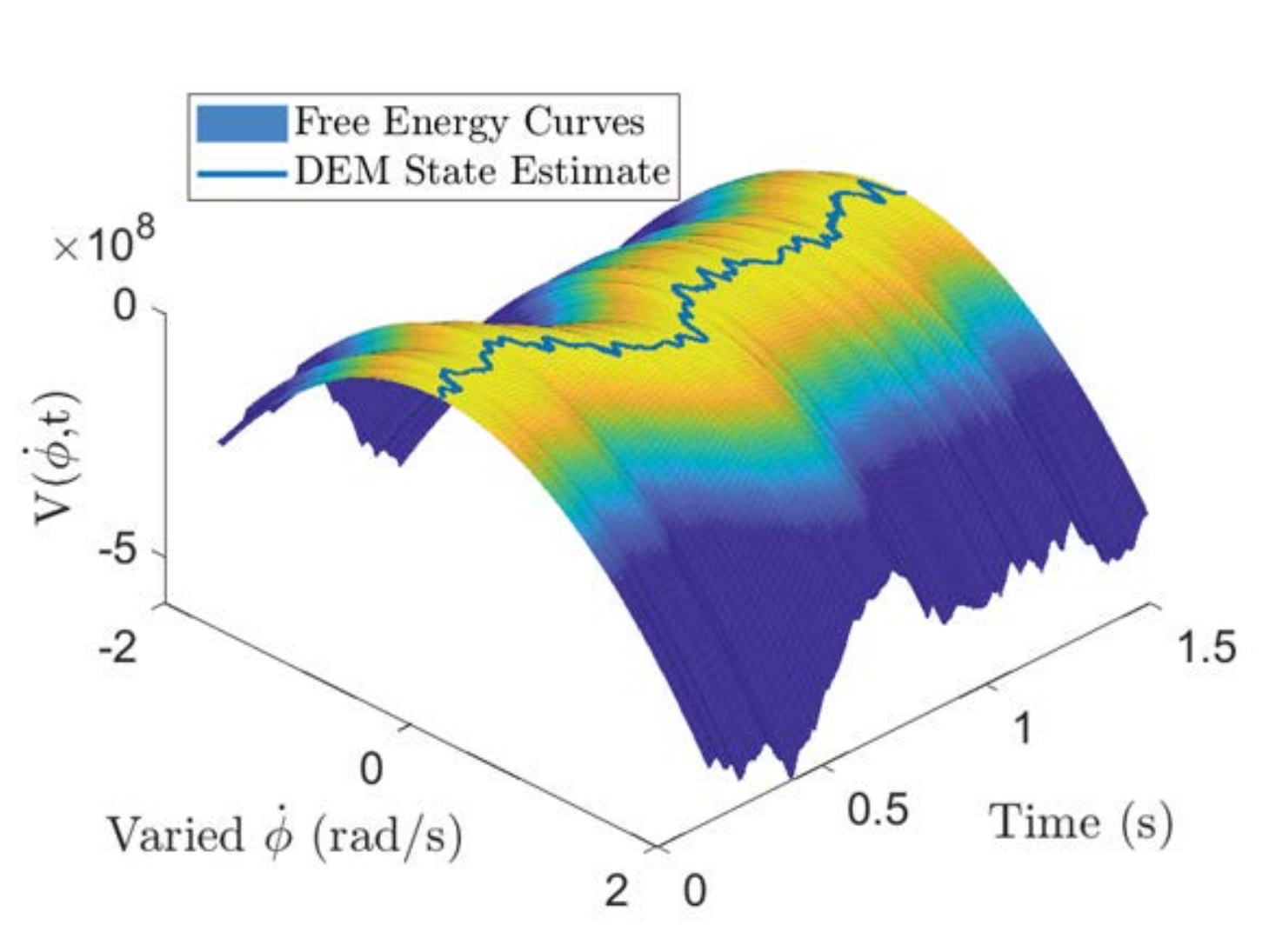}
    \caption{The DEM state estimate (blue curve) lies on top of the variational free energy surface, indicating that the DEM observer maximized $V(t)$.}
    \label{fig:Free_energy_curve}
\end{figure}

The fundamental idea behind state estimation using DEM is the gradient ascent over the variational free energy manifold. In this section, we visually demonstrate that DEM's state estimates for flight experiment maximize the VFE. Figure \ref{fig:Free_energy_curve} shows that the DEM state estimate is on top of the $V(t)$ curve at each time instance.

\begin{figure}[h]
    \centering
    \captionsetup{justification=justified}
    \includegraphics[scale = 0.4]{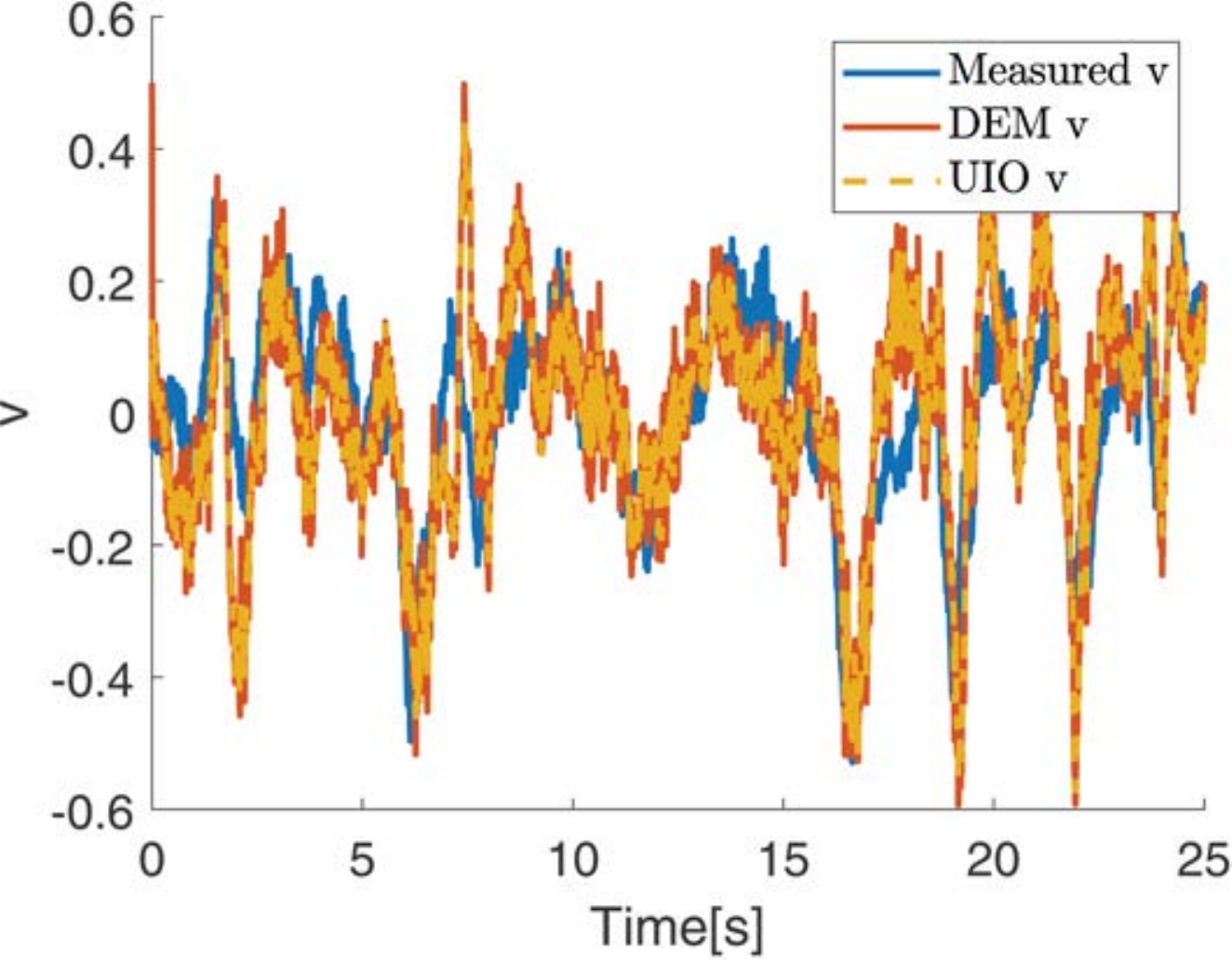}
    \caption{DEM's input estimation coinciding with that of UIO.}
    \label{fig:DEM_VS_UIO}
\end{figure}

\subsection{Input Estimation - benchmarking} 
In this section, we aim to demonstrate our DEM observer's capability to estimate inputs in real robot application and benchmark it against an input observer (UIO) from control systems.  We use the same settings as given in Section \ref{sec:settings}, except for providing the input priors for $pwm_1$ with a wrong value of $\eta^{pwm_1} = 0.5$ with a low precision of $P^{pwm_1} = 1$ to encourage exploration and $\Pi^w=e^3I_2$. We use $C=I$ for this section to meet the observability requirements of our benchmark (UIO). Both DEM and UIO estimated the first pwm signal and the result is shown in Figure \ref{fig:DEM_VS_UIO}. Both DEM and UIO followed the trend of measured inputs (in blue).

\begin{figure}[h]
    \centering
    \captionsetup{justification=justified}
    \includegraphics[scale = 0.3]{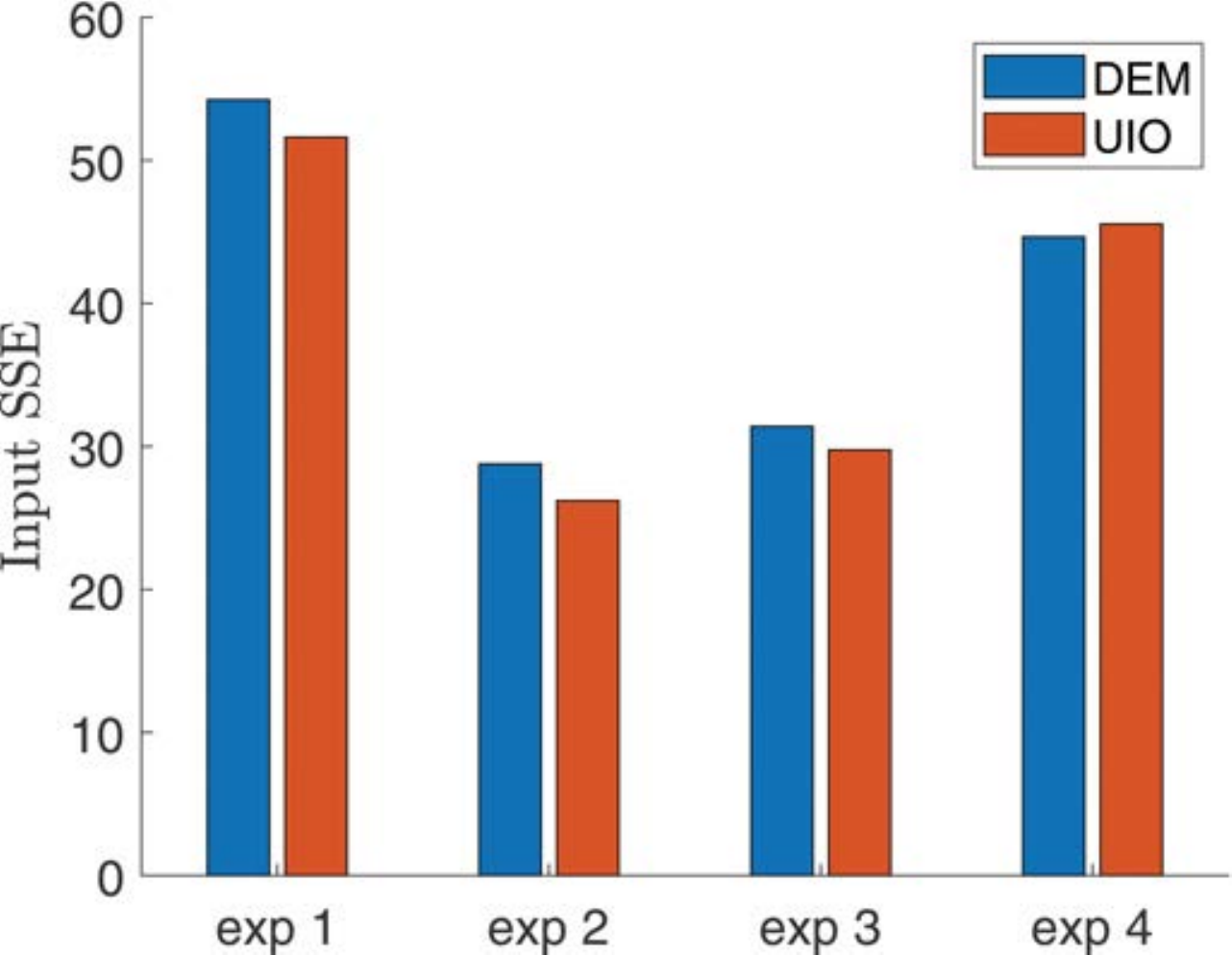}
    \caption{Similar performance of DEM and UIO for input estimation indicated by similar SSE in input estimation.}
    \label{fig:bar_UIO}
\end{figure}

The coinciding input estimates for DEM and UIO demonstrate that both estimators behave the same. The estimation was repeated for all experiments and the SSE for input estimation is shown in Figure \ref{fig:bar_UIO}. This confirms the similarity in performance of UIO and DEM for input estimation in the presence of colored noise.

% \textcolor{red}{We could shape this such that DEM can be used for fault detection. We provide our belief as priors and DEM finds the correct inputs.}
% \textcolor{red}{for input estimation Pw = eye(2) * exp(3) the rest is the same as section 5C. Only for figure \ref{fig:DEM_VS_UIO} and \ref{fig:bar_UIO} the C matrix has been changed to [1 0;0 1]. For accuracy vs complexity plot the settings are the same 5C apart from Pw which is exp(3)}

% Next to state estimation, DEM also performs input estimation. With the generative model and the sensory input, DEM is capable of estimating the least surprising inputs based on the given prior beliefs and the precisions, or certainties, on those priors. The capability of DEM input prediction is a usefull feature in real-life applications where inputs are uncertain or even unknown. In figure \ref{fig:input_est_vary} it can be observed how, when relaxing the prior precision, $P_v$, the DEM input estimate moves from the incorrect prior to the measured input. 

\begin{figure}[h]
    \centering
    \captionsetup{justification=justified}
    \includegraphics[scale = 0.4]{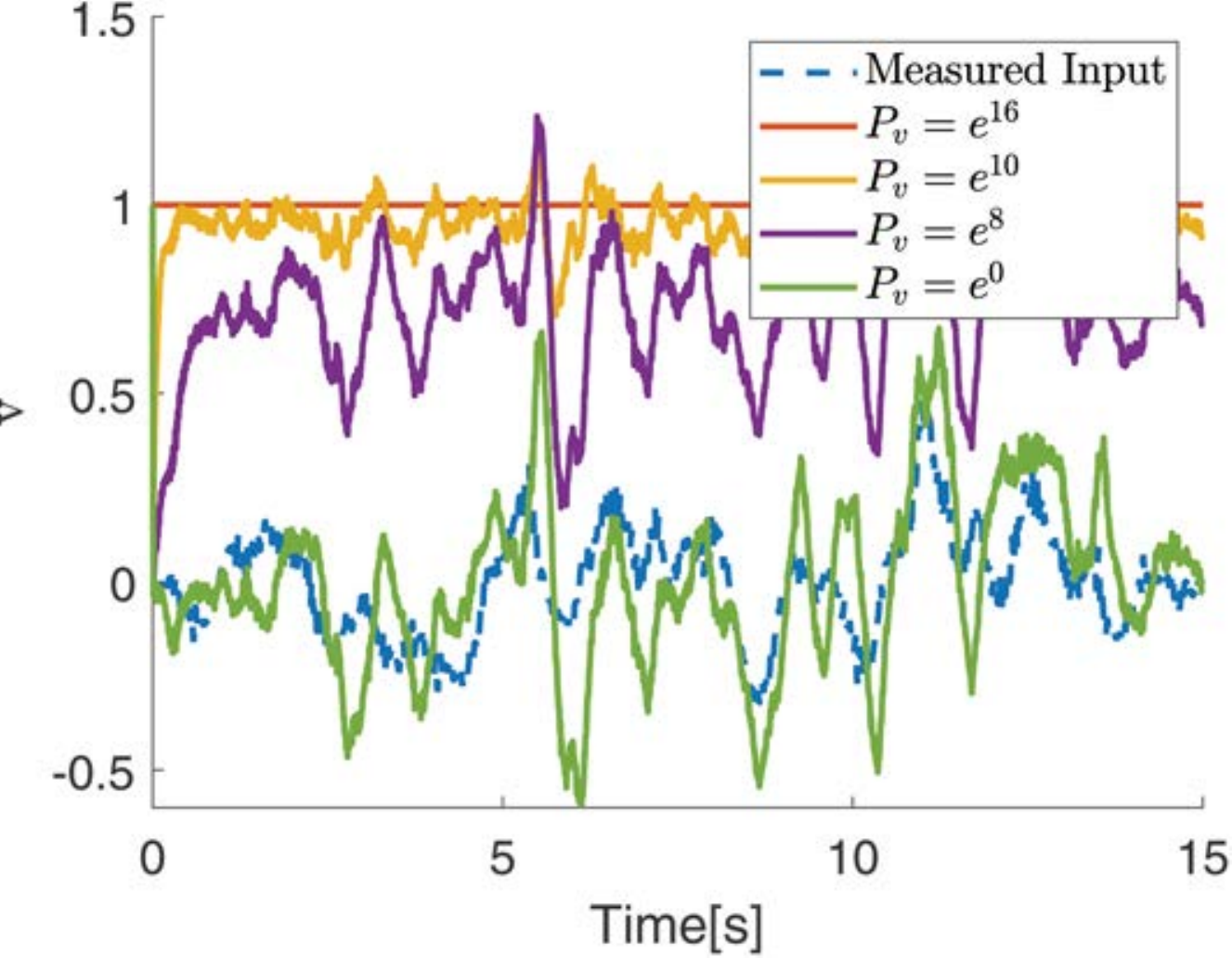}
    \caption{The input estimates moving from a wrong prior of $\eta^v =1$ to the measured input (in blue), mediated by the prior precision $P_v$.}
    \label{fig:input_est}
\end{figure}

\subsection{Accuracy v/s complexity}
\begin{figure}[h]
    \centering
    \captionsetup{justification=justified}
    \includegraphics[scale = 0.4]{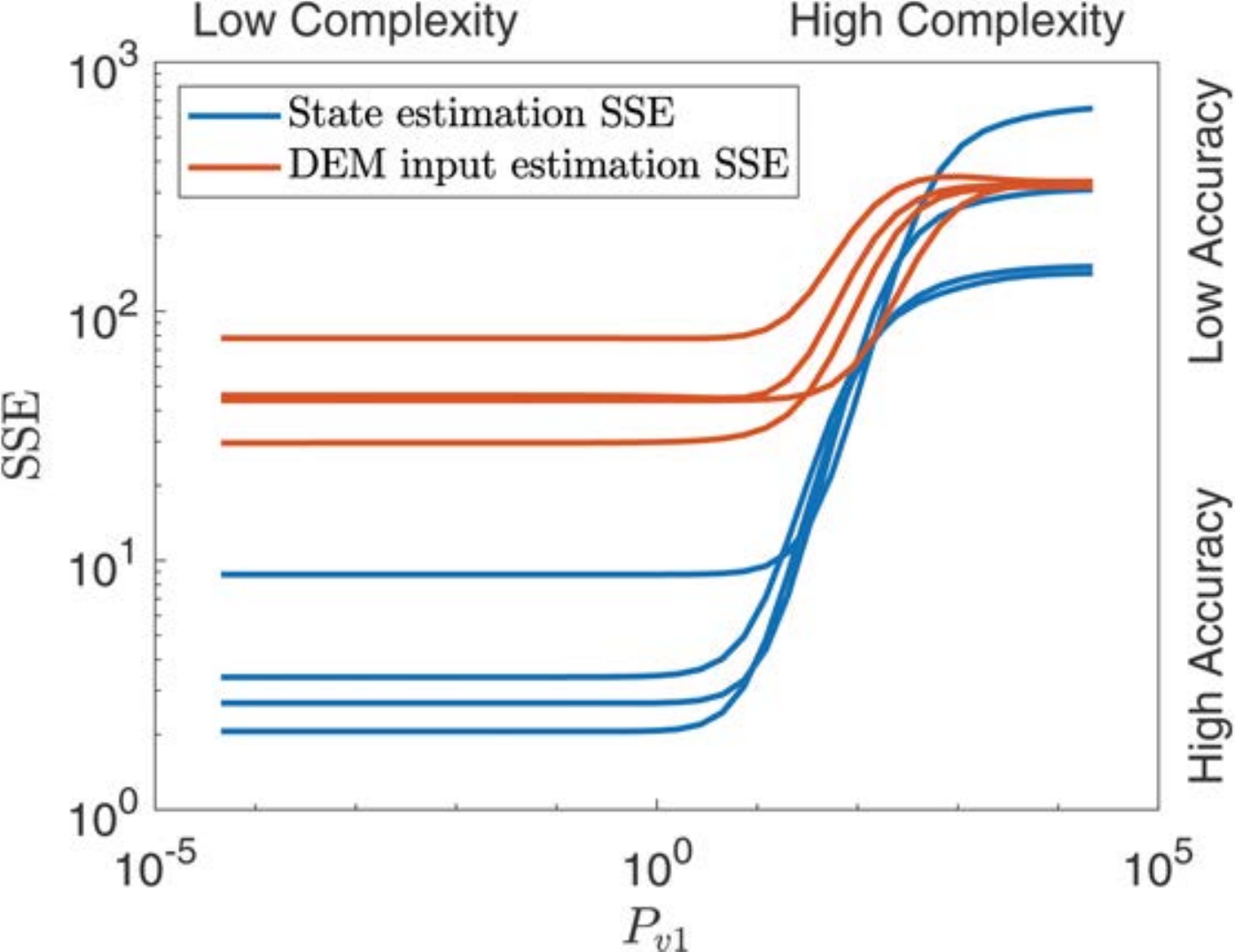}
    \caption{The SSE plot of state and input estimation demonstrating DEM's accuracy-complexity tradeoff. The SSE moves from a region of high complexity and low accuracy to a region of low complexity and high accuracy as the prior precision $P^v$ is relaxed. }
    \label{fig:Pv_vs_SSE}
\end{figure}

The inherent capability of DEM to balance between estimation accuracy and complexity is mediated by the priors $\eta^v$ and $P^v$. Here, accuracy is the measure of closeness of estimates to the real measurement, and complexity is the measure of deviation from the priors. This section aims at demonstrating this balance for simultaneous state and input estimation on quadrotor data. This section follows the same settings as Section \ref{sec:settings} with $\Pi^w=e^3I_2$. Simultaneous state and input estimation was performed using wrong input prior $\eta^v=1$ for varying prior precisions $P^v$, and the resulting input estimation is shown in Figure \ref{fig:input_est}, along with the measured input (in blue). As $P^v$ is relaxed, the input estimation moves away from the wrong prior $\eta^v$ and moves closer to the correct inputs. The shift from wrong priors to the correct measurements, mediated by $P^v$ can be seen as a balance (trade-off) between complexity and accuracy. Figure \ref{fig:Pv_vs_SSE} demonstrates this balance for all experiments with windy conditions, both for state and input estimation. The increasing SSE for higher $P^v$ indicates the shift from high accuracy with low complexity region to the low accuracy with high complexity region. This trade-off is useful mainly in industrial fault detection systems where any major deviations from the prior (known) inputs could be detected and isolated during runtime. DEM's inherent capability to balance accuracy and complexity is an added advantage when compared to other input estimators in literature like UIO.

% This can be observed in figure \ref{fig:Pv_vs_SSE} where the precision on the incorrect prior of input 1 is varied. For a high precision, the estimate has a high complexity, focusing on the prior, which leads to a low accuracy. When the prior is relaxed, both the input and state estimations of DEM increase in terms of accuracy.  

